\documentclass[11pt,a4paper]{article}
\usepackage[hyperref]{conll-2019}
\usepackage{times}
\usepackage{latexsym}

\usepackage{url}
\usepackage{multirow}

\usepackage{amsmath}
\usepackage{graphicx}
\usepackage{arabtex}
\usepackage[T1]{fontenc}
\usepackage{wrapfig}
\usepackage{enumitem}
\usepackage{colortbl,xcolor}
\usepackage{footnote}

\usepackage{tipa}

\newcommand{\hide}[1]{}

\newcommand{\SHADDA}{{$\sim$}}

\usepackage{float}

\aclfinalcopy 

\title{Joint Diacritization, Lemmatization, Normalization, and \\ Fine-Grained Morphological Tagging}

\author{Nasser Zalmout \and Nizar Habash \\
  Computational Approaches to Modeling Language Lab \\
  New York University Abu Dhabi \\
  United Arab Emirates \\
  {\tt \{nasser.zalmout,nizar.habash\}@nyu.edu} \\}

\date{}

\begin{document}

\maketitle
\setarab
\novocalize

\begin{abstract}
Semitic languages can be highly ambiguous, having several interpretations of the same surface forms, and morphologically rich, having many morphemes that realize several morphological features. This is further exacerbated for dialectal content, which is more prone to noise and lacks a standard orthography. The morphological features can be lexicalized, like lemmas and diacritized forms, or non-lexicalized, like gender, number, and part-of-speech tags, among others. 
Joint modeling of the lexicalized and non-lexicalized features can identify more intricate morphological patterns, which provide better context modeling, and further disambiguate ambiguous lexical choices. 
However, the different modeling granularity can make joint modeling more difficult.
Our approach models the different features jointly, whether lexicalized (on the character-level), where we also model surface form normalization, or non-lexicalized (on the word-level). We use Arabic as a test case, and achieve state-of-the-art results for Modern Standard Arabic, with 20\% relative error reduction, and Egyptian Arabic (a dialectal variant of Arabic), with 11\% reduction.

\end{list}
\end{abstract}

\section{Introduction}
Morphological modeling in Semitic languages can be very challenging. The optional short vowels (diacritics), along with the different possible interpretations of the same words, increase the ambiguity of surface forms. Moreover, the morphological richness of these languages results in large target spaces, which increase the model sparsity.
The different morphological features can be modeled through combined feature tags, using a single (but very large) target space, or through having separate models for each of the features.  The combined features approach models the relationships between the different features explicitly, but the large target spaces for morphologically rich languages further increase sparsity. On the other hand, separate feature modeling guarantees smaller target spaces for the individual features, but the hard separation between the features prevents modeling any inter-feature dependencies, and constrains the overall modeling capacity.

The set of morphological features can include lexicalized and non-lexicalized features, which further exacerbates joint modeling. Non-lexicalized features, like gender, and number, among others, have limited target spaces, and usually modeled as classification tasks. Lexicalized features, like lemmas and diacritized forms (for Semitic languages), are open-ended, with large target vocabularies. Moreover, non-lexicalized features are modeled on the word-level, whereas lexicalized features are optimally modeled on the character-level. This difference in the modeling granularity can be challenging for joint models. 

In this paper we present a model for handling lexicalized and non-lexicalized features jointly. We use a sequence-to-sequence based architecture, with different parameter sharing strategies at the encoder and decoder sides for the different features. The non-lexicalized features are handled with a tagger, which shares several parameters with the encoder, and uses a multitask-learning architecture to model the different non-lexicalized features jointly. The lexicalized features, on the other hand, are handled with a specific decoder for each feature, sharing the same encoder. Our architecture models the non-lexicalized features on the word-level, with a context-representation that spans the entire sentence. The lexicalized features are modeled on the character-level, with a fixed character context window around the target word. 
The character-level modeling is also suitable for surface form normalization, which is important for noisy content that is common in dialectal variants.

We use Modern Standard Arabic (MSA) and Egyptian Arabic (\textsc{Egy}) as test cases. Traditional disambiguation approaches for Arabic model each of the features separately, followed by an explicit ranking step using an external morphological analyzer.  Our joint model achieves 20\% relative error reduction (1.9\% absolute improvement) for MSA, and 11\% relative error reduction (2.5\% absolute improvement) for \textsc{Egy}, compared to the baseline that models the features separately.

\section{Background and Related Work}
In this section we present a brief linguistic overview of the challenges facing morphological modeling in the Semitic and morphologically rich languages. We then discuss related contributions in literature, and how our model compares to them.

\subsection{Linguistic Introduction}

\begin{table*}[hbt!]
\setlength{\tabcolsep}{3pt}
\resizebox{1.0\textwidth}{!}{%
\begin{tabular}{ l  l  l  l  c  c  c  c  c  c  c  c  c  c  c  c  c  }
\hline
\bf Diacrtization & \bf Lemma & \bf English & \bf POS & \bf Prc3 & \bf Prc2 & \bf Prc1 & \bf Prc0 & \bf Per & \bf Asp & \bf Vox & \bf Mod & \bf Gen & \bf Num & \bf Stt & \bf Cas & \bf Enc0 \\ \hline
lam{\SHADDA}atohum & lam{\SHADDA} & she collected them  & verb  & 0  & 0  & 0  & 0  & 3  & p  & a  & i  & f  & s  & na  & na  & dobj\textsubscript{3mp} \\ \hline
lumotahum & lAm & you [m.s.] blamed them  & verb  & 0  & 0  & 0  & 0  & 2  & p  & a  & i  & m  & s  & na  & na  & dobj\textsubscript{3mp} \\ \hline
lumotihim & lAm & you [f.s.] blamed them  & verb  & 0  & 0  & 0  & 0  & 2  & p  & a  & i  & f  & s  & na  & na  & dobj\textsubscript{3mp} \\ \hline
lumotuhum & lAm & I blamed them & verb  & 0  & 0  & 0  & 0  & 1  & p  & a  & i  & m  & s  & na  & na  & dobj\textsubscript{3mp} \\ \hline
lam{\SHADDA}atuhum & lam{\SHADDA}ap & their collection & noun  & 0  & 0  & 0  & 0  & na  & na  & na  & na  & f  & s  & c  & n  & poss\textsubscript{3mp} \\ \hline
limut{\SHADDA}ahamK & mut{\SHADDA}aham & for a suspect & noun  & 0  & 0  & li (prep)  & 0  & na  & na  & na  & na  & m  & s  & i  & g  & 0 \\ \hline
limut{\SHADDA}ahimK & mut{\SHADDA}ahim & for an accuser & noun  & 0  & 0  & li (prep)  & 0  & na  & na  & na  & na  & m  & s  & i  & g  & 0 \\ \hline

\end{tabular}%
}
\caption{A subset of all the possible analyses for the word <lmthm> {\it lmthm}. Notice that in the last two analyses the words are disambiguated through the lemmas and diacritized forms only, and they share all the other features.}
\label{joint_modeling_example_table}
\end{table*}

Morphologically rich languages (MRLs) tend to have more fully inflected words than other languages, realized through many morphemes that represent several morphological features. The target space for the combined morphological features therefore tends to be large, which increases sparsity. MRLs also can be highly ambiguous, with different interpretations of the same surface forms. Ambiguity is further exacerbated for Semitic languages, like Arabic and Hebrew, at which the short vowels (diacritics) can be kept or dropped. The high degree of ambiguity in Arabic results in having about 12 analyses per word on average \cite{Habash_10_book}.

Both morphological richness and ambiguity can be modeled with \textit{morphological analyzers}, or morphological dictionaries, which are used to encode all potential word inflections in the language. Morphological analyzers should ideally return all the possible analyses of a surface word (to model ambiguity), and cover all the inflected forms of a word lemma (to model morphological richness), covering all related features. The best analysis can then be chosen through \textit{morphological disambiguation}; by predicting the different morphological feature values and use them to rank the relevant analyses from the analyzer. The morphological features that we model for Arabic include:

\begin{itemize}
\item Lexicalized features: lemmas (lex) and diacritized forms (diac).
\item Non-lexicalized features: aspect (asp), case (cas), gender (gen), person (per), part-of-speech (POS), number (num), mood (mod), state (stt), voice (vox).
\item Clitics: enclitics, like pronominal enclitics, negative particle enclitics; proclitics, like article proclitic, preposition proclitics, conjunction proclitics, question proclitics.
\end{itemize}

Table \ref{joint_modeling_example_table} shows an example highlighting the different morphological features. The example presents a subset of the possible analyses for the word  <lmthm> {\it lmthm}. The different features can be used to disambiguate the right analysis. Disambiguation using the non-lexicalized features only might not be conclusive enough, as we see in the last two analyses, where the lemma and diacritized form only can disambiguate the right analysis.

Dialectal Arabic (DA) includes several dialects of Arabic, like \textsc{Egy}, that vary by the geographical location in the Arab world. DA is also Semitic and an MRL, but it is mainly spoken, and lacks a standard orthography \cite{CODA:TR:2012}. The lack of a standard orthography further increases sparsity and ambiguity, hence requiring explicit normalization. \newcite{CODA:TR:2012,habash-etal-2018-unified} proposed CODA, a Conventional Orthography for Dialectal Arabic, which aims to provide a conventionalized orthography across the various Arabic dialects. We use this convention as the reference for the normalization task.

\subsection{Morphological Tagging}

Arabic morphological tagging and disambiguation have been studied extensively in literature, with contributions for MSA \cite{khalifayamama,FARASA:2016NAACL,ACL:habash-rambow:2005,Diab:2004a}, and DA \cite{Habash_13_NAACL,konvens:09_alsabbagh12o,Duh:2005}. There are also several recent contributions that showed significant accuracy improvement using deep learning models  \cite{Zalmout_2018_NAACL,go:2017,Zalmout-2017-EMNLP,14-languages-tagging}. In addition to other deep learning contributions that showed limited success for Arabic \cite{shen-EtAl:2016:COLING1}.
Most of these contributions model the different morphological features separately, or focus on a limited feature subset. We elaborate on the contributions with some joint modeling aspects later in the section.

\subsection{Diacritization and Lemmatization}
Diacritization and lemmatization are very useful for tasks like information retrieval, machine translation, and text-to-speech, among others.

Diacritization has generally been an active area of research \cite{darwish2017arabic,zitouni-sorensen-sarikaya:2006:COLACL,ACL:nelken-shieber:2005}. More recent contributions use Deep Learning models in different configurations;  \newcite{belinkov-glass:2015:EMNLP} model diacritization as a classification task, using  Long Short Term Memory (LSTM) cells. And \newcite{abandah2015automatic} use LSTMs to model diacritization as a sequence transcription task, similar to \newcite{mubarak-etal-2019-highly} who model diacritization as a sequence-to-sequence task. 

Early contributions for lemmatization used finite state machines \cite{schmid2004smor,minnen_carroll_pearce_2001}, which had a limited capacity for modeling unseen words or lemmas. There were also several contributions that utilize a joint tagging and lemmatization approach, using CRFs and Maximum Entropy models \cite{muller2015joint,chrupala2008learning}.
Other contributions approached lemmatization as a lemma-selection task \cite{ezeiza1998combining}, where the goal is to select the correct lemma from a set of lemmas  provided by a morphological analyzer. Many of the lemmatization models for Arabic use a similar approach \cite{MADAMIRA:2014,RothEtAl:2008:ShortPapers}. More recently, sequence-to-sequence models with attention \cite{bahdanau2014neural} have been shown useful in several NLP tasks, with several lemmatization contributions \cite{DBLP:journals/corr/abs-1904-02306,bergmanis2018context,putz2018seq2seq}. Other contributions use additional morphosyntactic features as part of the modeling architecture \cite{kanerva2019universal,kondratyuk2018lemmatag}, somewhat similar to our approach.

\subsection{Joint Morphological Modeling in Arabic}

There are also several contributions for the joint modeling of the different morphological features in Arabic. However, most of these contributions use separate models for each of the features, and usually use a ranking step to select the best overall morphological analysis from an external morphological analyzer \cite{RothEtAl:2008:ShortPapers,habash-rambow:2007:ShortPapers}. MADAMIRA \cite{MADAMIRA:2014} is a popular system for Arabic morphological tagging and disambiguation. It uses SVMs for the different non-lexicalized features, and n-gram language models for the lemmas and diacritized forms. \newcite{Zalmout-2017-EMNLP} presented a neural extension of this model, with LSTM taggers for the individual features, and neural language models for the lexicalized features. \newcite{go:2017} used multi-task learning for fine-grained part-of-speech tagging, modeling the different morphological features jointly, but they do not model lemmas or diacritized forms. \newcite{Zalmout-ACL-2019} also used multitask learning for the different non-lexicalized morphological features, and neural language models for lemmas and diacritized forms. This model currently provides state-of-the-art results for Arabic.

\paragraph{Surface Form Normalization} In the joint morphological models that rely on morphological analyzers \cite{Zalmout-ACL-2019,Zalmout-2017-EMNLP,MADAMIRA:2014} surface form normalization (through presenting the words in CODA form) are byproducts of selecting the correct analysis, rather than being explicitly modeled.

\section{Approach}
Non-lexicalized features are usually modeled on the word-level, whereas lexicalized features are better handled through character-level models. Moreover, the context representation for morphological tagging of the non-lexicalized features usually spans the entire sentence, using LSTMs for example. The optimal context representation for the lexicalized features, on the other hand, is through a fixed number of characters before and after the target word \cite{bergmanis2018context}. This difference in modeling granularity, whether in terms of context representation or word/character level modeling, can be very challenging for a joint modeling approach. 

We use a modified sequence-to-sequence architecture, where some components of the encoder are shared between a tagger, for the non-lexicalized features, and the encoder-decoder architecture, for the lexicalized features. We also use separate decoders for the different lexicalized features, that share the same encoder and trained jointly using a shared loss function. The remainder of this section discusses the architecture in more detail.

\subsection{Encoder}

We use two Bi-LSTM layers for the hidden representation at the encoder. The input context is modeled through a sliding window of a fixed number of characters around the target word, as in the Lematus model \cite{bergmanis2018context}. We also use additional special symbols for the whitespace and target word boundaries. In addition to the character embeddings, we also condition on the word-level embedding of the word containing the characters. We concatenate the word embedding vector with the input character embeddings. Each character embedding $\mathbf{c}_i$ is replaced by the concatenation $\left[\mathbf{c}_i; \mathbf{w}_j\right]$ before being fed to the encoder, where $\mathbf{w}_j$ is the $d_{w}$-dimensional word embedding of the word $j$ in which character $i$ appears in. Given the characters of input sentence $c$ and its lemmatized equivalent $y$, the goal is to model $P(y_k |\mathbf{c}_i,\mathbf{w}_j)$.

Several previous contributions for Arabic showed that pretraining the word embeddings is very useful  \cite{erdmann-etal-2018-addressing,watson2018utilizing,Zalmout-2017-EMNLP}, including the baselines used in this paper. We therefore pre-train the word embeddings with FastText \cite{bojanowski2017enriching}, using a large external dataset. Whereas the character embeddings are learnt within the model.

\subsection{Tagger}

The tagging architecture is similar to the architecture presented by \newcite{Zalmout-ACL-2019}, but we share the character and word embeddings from the encoder with the tagger network, used for the non-lexicalized features. We use two Bi-LSTM layers on the word-level to model the context for each direction of the target word. The context in the tagging network spans the entire input sentence, rather than a fixed window as in the encoder.  For each sentence of length L  \(\{w_1, w_2, ..., w_L\)\}, every word \(w_j\) is represented by vector $\mathbf{v}_j$, which is comprised of the concatenation:

\[\mathbf{v}_j = [\mathbf{w}_j; \mathbf{s}_j; \mathbf{a}_j]\]

Where $\mathbf{w}_j$ is the word embedding vector, \(\mathbf{s}_j\) is a vector representation of the characters within the word, and \(\mathbf{a}_j\) is a vector representing all the candidate morphological tags (from an analyzer), for all the non-lexicalized morphological features.

To obtain the vector \(\mathbf{s}_j\), we use an LSTM-based model, applied to the character sequence in each word separately. We use the last state vector as the embedding representation of the word's characters. 
Whereas to get the \(\mathbf{a}_j\) vector, for each morphological feature $f$, we use a morphological analyzer to obtain all possible feature values of the word to be analyzed. We then embed each value separately (with separate embedding tensors for each feature, learnt within the model), then sum all the resulting vectors to to get  \(\mathbf{a}_j^{f}\) (since these tags are alternatives and do not constitute a sequence). We concatenate the individual  \(\mathbf{a}_j^{f}\)  vectors for each morphological feature \(f\) of each word, to get a single representation, \(\mathbf{a}_j\), for all the features:
\begin{align*}
& \mathbf{a}_j^{f} = \sum_{n=1}^{N_f} \mathbf{a}_{j,n}^{f} \\
& \mathbf{a}_j = [\mathbf{a}_j^{pos};...;\mathbf{a}_j^{num};...;\mathbf{a}_j^{vox}]
\end{align*}
Where \(N_f\) is the set of possible candidate values for each feature \(f\) (from the analyzer).
The \(\mathbf{a}_j\) vector does not constitute a hard constraint and can be discarded if a morphological analyzer is not used.

\begin{figure}[h]
\centerline{%
\includegraphics[scale=0.45]{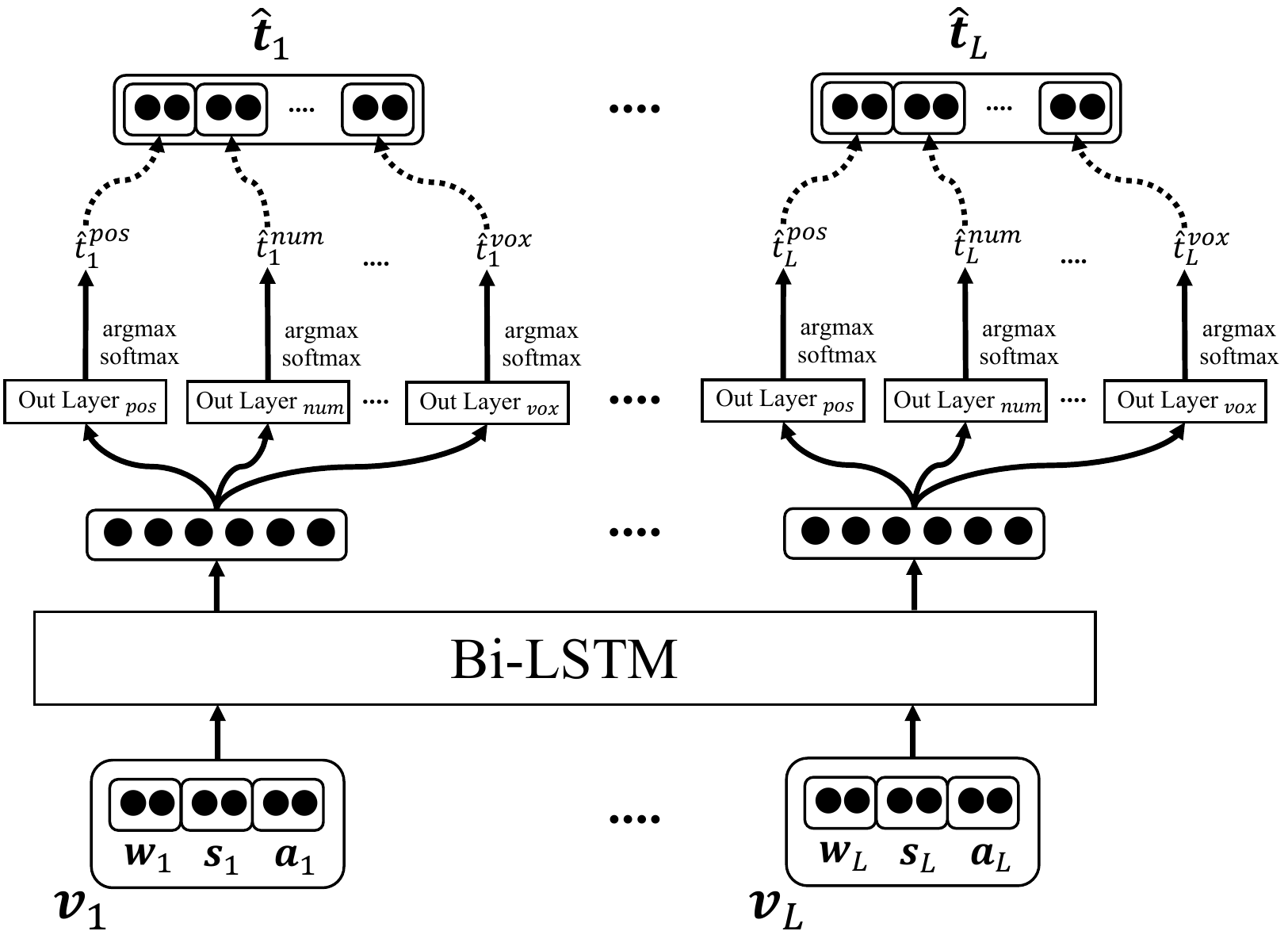}
}%
\caption{The tagger model, showing the multitask learning architecture for the features. The concatenated predicted tags are used to condition on, at the decoders.}
\label{tagger_architecture_concatenated_tags}
\end{figure}

We use a multitask learning setup to train the different morphological features jointly, through sharing the parameters of the hidden layers in the Bi-LSTM network. The input is also shared, through the $\mathbf{v}_j$ vector.
The output of the network is then fed to a separate non-linearity function, output layer, and softmax, for a probability distribution of each of the features separately. Figure \ref{tagger_architecture_concatenated_tags} shows the overall tagging architecture. 

\subsection{Decoders}

We use separate decoders for lemmatization and diacritization, with two LSTM layers for each. Both decoders share the same input and parameters of the encoder Bi-LSTM network. 
For each decoder, we condition on the decoder output of the previous step, along with Luong attention \cite{luong2015effective} over the encoder outputs $h_i$, and the predicted tags from the tagger. We use the last encoder output as the initial states for the decoder layers. We use scheduled sampling \cite{bengio2015scheduled} during training, and feed the $d_{c}$-dimensional character embeddings at every time step. But we found empirically that using a constant sampling probability instead of scheduling provides better results. For both the encoder and decoder RNNs, we also use dropout on the non-recurrent connections of both the encoder and decoder layers during training.
The decoder outputs are fed to a softmax layer that reshapes the vectors to dimension $d_{voc}$, then argmax to yield an output sequence $\vec{y}$ one character at a time. 

\paragraph{Conditioning on the Predicted Tags} In addition to the attention distribution and the previous time step, we also condition on the predicted tags from the tagger during decoding. The goal is to provide an additional contextual signal to the decoders, and to disambiguate the possible lexical choices. We use the output of the argmax (over the softmax distribution) for each feature, and concatenate the different tags in a similar representation to the \(\mathbf{a}_j\) vector: 

\[ \hat{\mathbf{t}}_j = [\hat{\mathbf{t}}_j^{asp};...;\hat{\mathbf{t}}_j^{pos};...;\hat{\mathbf{t}}_j^{vox}] \]

\paragraph{Preventing Backpropagation from the Decoders to the Tagger} The decoder produces the lexicalized features at the character-level, whereas the predicted tags are on the word-level. We found that Backpropagating the gradients from the decoder to the tagger network leads to instability at the tagger, and we thought that the different granularities might create some biases. Therefore, we prevent the decoder from backpropagating gradients to the tagger during training. This is consistent with the architecture presented by \newcite{kondratyuk2018lemmatag}.

\begin{figure}[h]
\centerline{%
\includegraphics[scale=0.4]{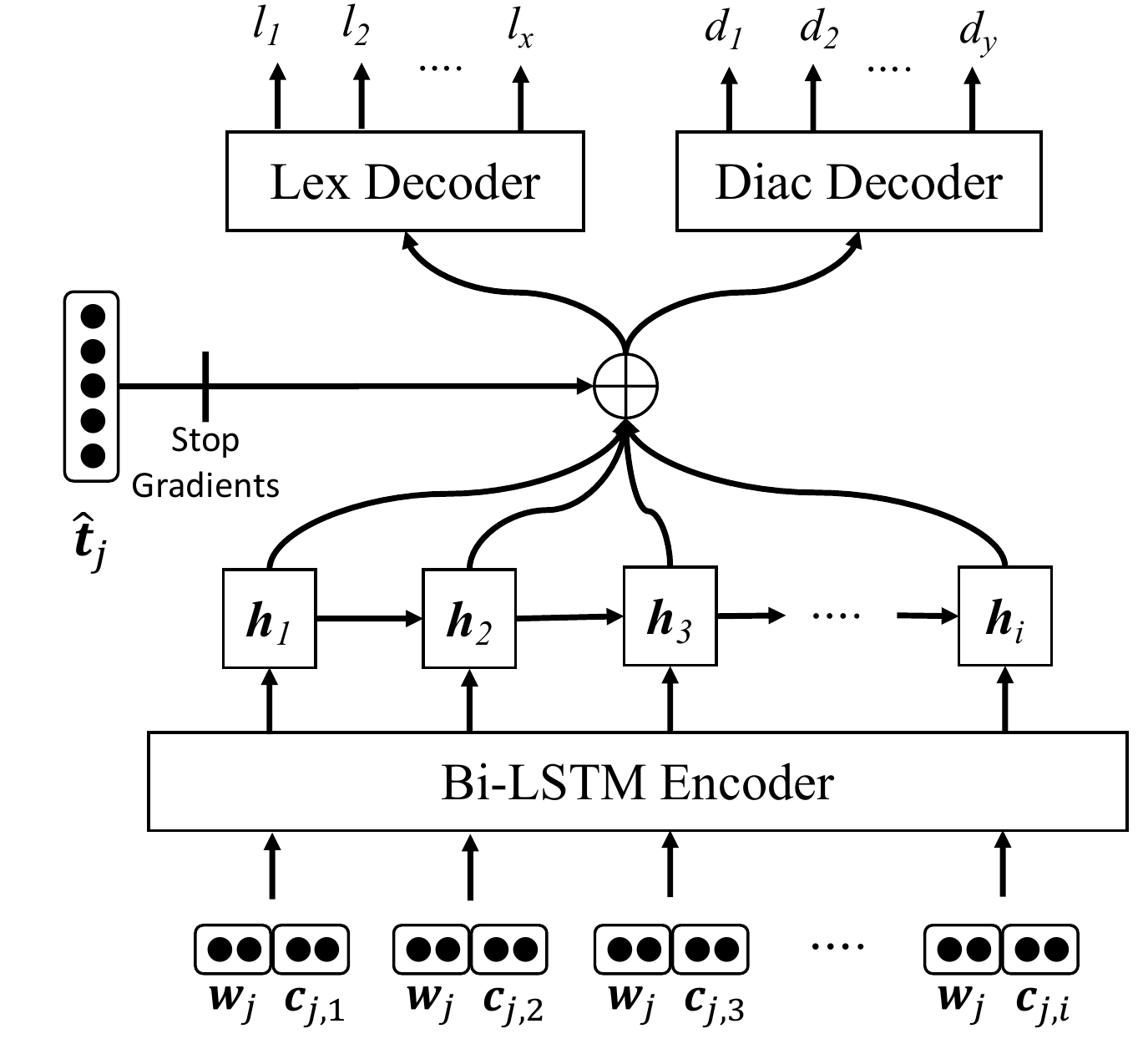}
}%
\caption{The sequence-to-sequence architecture for the lexicalized features, with a shared encoder, and separate decoders for lemmatization and diacritization. The figure does not show the fixed context window of 10 characters before and after the target word.}
\label{mtl_decoders}
\end{figure}

\subsection{Surface Form Normalization} 

The normalization task is particularly important for dialectal content, which lack a standardized orthography. The training data that we use has the diacritized annotations already in the CODA normalized form for \textsc{Egy}. So the output sequence of the diacritization task should be both the diacritized and CODA normalized version of the input sequence. This normalization is learnt explicitly in our character-level sequence-to-sequence model.
For MSA there is no need for CODA normalization, so the normalized output includes any error correction that might happen in the training dataset. Normalization is assessed as part of the overall diacritization accuracy.

\subsection{Training Procedure}
We use a small held-out tuning set of about 5\% of the training data to save the best model during training. We did not use the development set here to be consistent with other contributions in literature, where the development set is primarily used to evaluate high level design decisions only. We train the model for a fixed number of epochs and select the model that performs best on the tuning set. This method provided the most stable results, compared to early stopping or other methods.

The loss function is based on minimizing cross entropy \(H\) for each feature \(f\). The overall loss is the average of the individual losses for the different features, whether lexicalized or non-lexicalized:

\[ H(\hat{y}, y) = \frac{1}{|F|}\sum_{f \in F}  H(\hat{y}^f, y^f)  \]

Where \(F\) is the set of features that we model. \(y\) represents the true feature value, and \(\hat{y}\) is the predicted value. 
We experimented with having different optimizers for the lexicalized and non-lexicalized features. We also experimented with a weighted average for the different features, where the weights are learnt as part of the end-to-end system. None of these modifications provided any improvement.
We use Adam optimizer \cite{Kingma-arxiv-2014} with a learning rate of 0.0005, and we run the various models for 50 epochs.

\subsection{Full Morphological Disambiguation}

Morphological disambiguation involves predicting the right combination of morphological features for each word in context. We can either present the predicted features from the model directly, or use a morphological analyzer to guarantee more consistent feature values. If a morphological analyzer is used, the disambiguation system selects the optimal analysis for the word from the set of analyses returned by the analyzer. We use the predicted tags to rank the analyses, and select the analysis with highest number of matched feature values. The different features can be assigned different weights during ranking. Refer to other contributions in literature that use a similar approach for more details  \cite{Zalmout-ACL-2019,Zalmout-2017-EMNLP,MADAMIRA:2014}.

\section{Experiments and Results}
\subsection{Data}

We use the Penn Arabic Treebank (PATB parts 1,2, and 3) \cite{Maamouri:2004} for \textsc{MSA}, and the ARZ dataset \cite{ARZMAG:2012} from the Linguistic Data Consortium (LDC), parts 1--5, for \textsc{Egy}. We follow the data splits recommended by \newcite{diab2013ldc} for \textsc{Train}, \textsc{DevTest}, and \textsc{BlindTest}. We use Alif/Ya and Hamza normalization. Both datasets include gold annotations for the diacritized forms, lemmas, and the remaining 14 features. The diacritized forms are normalized following the CODA guidelines for \textsc{Egy}.

Table~\ref{dataset} shows the data sizes. The \textsc{Tune} dataset is used during the model training process, for early stopping or to keep the best performing model. \textsc{Tune} is extracted randomly from the original \textsc{Train} split (almost 5\% of \textsc{Train}), so the other splits are consistent with the splits used in literature.  The \textsc{DevTest} dataset is used during the system development to assess design choices. The \textsc{BlindTest} dataset is used to evaluate the system after finalizing the architecture design, and to report the overall performance. 

\begin{table}[h]
\centering
\setlength{\tabcolsep}{5pt}
\begin{footnotesize}
\begin{tabular}{|c|c|c|c|c|}
\hline
 & \textbf{\textsc{Train}} & \textbf{\textsc{Tune}} & \textbf{\textsc{DevTest}} & \textbf{\textsc{BlindTest}} \\ \hline \hline
\textbf{\textsc{MSA}} & 479K & 23K & 63K & 63K \\ \hline
\textbf{\textsc{Egy}} & 127K & 6K & 21K & 20K \\ \hline
\end{tabular}
\end{footnotesize}
\caption{Word count statistics for MSA and \textsc{Egy}.}
\label{dataset}
\end{table}

We use the same morphological analyzers that were used in MADAMIRA \cite{MADAMIRA:2014}, and the other baselines, for both MSA and \textsc{Egy}. For MSA we use SAMA \cite{SAMA31}, and the combination of SAMA, CALIMA \cite{CALIMA:2012}, and ADAM \cite{salloum2014adam} for \textsc{Egy}. We use the LDC's Gigaword corpus \cite{LDC:Gigaword-5} to pretrain the MSA word embeddings, and the BOLT Arabic Forum Discussions corpus \cite{BOLT:2018} for \textsc{Egy}, as used in the reported baselines. We preprocessed both datasets with Alif/Ya and Hamza normalization, as we did for the training dataset.

\subsection{Experimental Setup}

\paragraph{Tagger} We use a similar setup as used by \newcite{Zalmout-ACL-2019}. We use two Bi-LSTM hidden layers of size 800, and dropout probability of 0.4, with peephole connections. The LSTM character embedding architecture uses two LSTM layers of size 100, and embedding size 50. 
We use FastText \cite{bojanowski2017enriching} to pretrain the word embeddings, with embedding dimension of 250, and an embedding window of size two.

\paragraph{Encoder-Decoder} We use two LSTM layers of size 400 for both the encoder and decoder (bidirectional for the encoder), dropout value of 0.4, fixed sampling probability of 0.4 \cite{bengio2015scheduled}. We use the same word and character embeddings as the tagger. We use beam decoding with beam size of 5, and a context window of 10 characters before and after the target word.

\paragraph{Metrics} The evaluation metrics we use include:

\begin{itemize}
\item POS accuracy (\textsc{POS}): The accuracy of the POS tags, of a tagset comprised of 36 tags \cite{Habash_13_NAACL}.
\item Non-lexicalized morphological features accuracy (\textsc{Tags}): The accuracy of the combined 14 morphological features we model, excluding lemmas and diacritized forms.
\item Lemmatization accuracy (\textsc{Lemma}): The accuracy of the fully diacritized lemma.
\item Diacritized forms accuracy (\textsc{Diac}): The accuracy of the diacritized (and CODA normalized for \textsc{Egy}) form of the words.
\item Full Analysis Accuracy (\textsc{Full}): Accuracy over the full analysis -- the strictest metric.
\end{itemize}

\paragraph{Baselines}

The first baseline is MADAMIRA \cite{MADAMIRA:2014}, which is one of the most commonly used morphological disambiguation models for Arabic. We also use the model suggested by \newcite{Zalmout-2017-EMNLP}, which is based on a similar architecture, but uses LSTM taggers instead of the SVM models in MADAMIRA, and LSTM-based language models instead of the n-gram models. The last baseline uses a multitask learning architecture to model the different non-lexicalized features jointly, but neural language models for the lexicalized features \cite{Zalmout-ACL-2019}. We use the same feature weights during the disambiguation process as this baseline.

\begin{table*}[h!]
\centering
\setlength{\tabcolsep}{4pt}
\def\arraystretch{1.0}
\begin{footnotesize}

\begin{tabular}{|l|l|c|c|c|c|c|}
\hline
\multicolumn{2}{|c|}{\multirow{2}{*}{\textbf{Model}}} & \multicolumn{5}{c|}{\textbf{\textsc{DevTest}}}  \\ \cline{3-7} 
\multicolumn{2}{|c|}{} & \textsc{Full} & \textsc{Tags} & \textsc{Diac} & \textsc{Lex} & \textsc{POS}  \\ \hline \hline
\multirow{5}{*}{\textbf{MSA}} & MADAMIRA (SVM models + analyzer) \cite{MADAMIRA:2014}& 85.6 & 87.1 & 87.7 & 96.3 & 97.1 \\ \cline{2-7} 
 & LSTM models + analyzer \cite{Zalmout-2017-EMNLP}  & 90.4 & 92.3 & 92.4 & 96.9 & 97.9 \\ \cline{2-7} 
 & \verb| |+ Multitask learning for the tags \cite{Zalmout-ACL-2019} & 90.8 & 92.7 & 92.7 & 96.9 & 97.9 \\ \cline{2-7} 
 & Joint modeling + analyzer & \textbf{92.3} & \textbf{93.5} & \textbf{93.9} & \textbf{97.6} & \textbf{98.1}  \\ \cline{2-7} 
 & Joint modeling without analyzer & 90.3 & 92.7 & 92.8 & 96.3 & 97.7 \\ \hline \hline
\multirow{5}{*}{\textbf{\textsc{Egy}}} & MADAMIRA (SVM models + analyzer) \cite{MADAMIRA:2014}  & 76.2 & 86.7 & 82.4 & 86.4 & 91.7  \\ \cline{2-7} 
 & LSTM models + analyzer \cite{Zalmout-2017-EMNLP}  & 77.0 & 88.8 & 82.9 & 87.6 & 92.9 \\ \cline{2-7} 
 & \verb| |+ Multitask learning for the tags \cite{Zalmout-ACL-2019} & 77.2 & 88.8 & 82.9 & 87.6 & \textbf{93.1}  \\ \cline{2-7} 
 & Joint modeling + analyzer & \textbf{79.5} & \textbf{89.0} & \textbf{85.0} & \textbf{88.5} & \textbf{93.1} \\ \cline{2-7} 
 & Joint modeling without analyzer & 73.2 & 84.9 & 81.5 & 84.4 & 91.1  \\ \hline
\end{tabular}

\end{footnotesize}
\caption{The results of the various models. The first and second baselines use separate models for the features, and the third uses a multitask learning architecture for the non-lexicalized features only. }
\label{joint_modeling_results}
\end{table*}

\subsection{Results}

Table \ref{joint_modeling_results} presents the results for the various baselines, and the results of joint modeling architecture. The results show a significant accuracy improvement for the joint modeling approach, compared to all baselines. 

\paragraph{Diacritization and Normalization} The diacritization task, which also includes surface form normalization, seems to have benefited the most of the joint modeling architecture, with about 16\% relative error reduction for MSA, and 12\% relative error reduction for \textsc{Egy}. This is probably due to the relatively large target space for diacritized forms when using the language modeling approach in the baseline, compared to lemmatization for example, which has a smaller overall types count. The character-level sequence-to-sequence architecture is more suitable to this task, with a small character target space.
Moreover, diacritization implicitly involves normalization based on our approach. Whereas in the baseline normalization is a byproduct of selecting the right analysis, rather than a modeling goal. Character-level models provide for an explicit and direct normalization capability, as the model learns to map the erroneous sequence to the normalized target sequence.

\paragraph{Overall Feature Consistency} An analysis is consistent if all the feature values are linguistically acceptable to co-occur with each other. For example, case is undefined for verbs, so if a verb analysis had a defined case value, this analysis is inconsistent. The same applies to consistency between the tags and the corresponding lemma (or diacritized form).
The \textsc{Tags} metric, which represents the accuracy of the combined non-lexicalized features, also shows noticeable improvement for MSA. The fact that \textsc{Tags} improved, along with \textsc{Full}, while the POS accuracy remained somewhat similar, indicates that the model is now producing more consistent morphological predictions. This improved consistency is probably the result of enhanced diacritization and lemmatization models, which provide a better signal to the overall analysis ranking. The improvement in \textsc{Tags} for \textsc{Egy}, on the other hand, is limited. This indicates that the model was probably already producing more consistent non-lexicalized morphological features, and the improvement in the \textsc{Full} metric is due to improved diacritization and lemmatization only.

\paragraph{The Role of Morphological Analyzers} Morphological analyzers are also used to guarantee consistency in the predicted features. The baselines and our best performing model all use morphological analyzers, to get the candidate tags at the input, and to produce the best analysis through the ranking process. We train our model without using the analyzer, to evaluate its role in the morphological disambiguation task. The results are lower, both for MSA and \textsc{Egy}. However, the result for MSA is very close to the \cite{Zalmout-2017-EMNLP} baseline, which uses separate feature models (with the analyzer). This indicates that our model can match the accuracy of a strong baseline, without relying on expensive external resources. This does not apply to \textsc{Egy}, probably due to the lower training data size and noisier content.

The accuracy gap between the systems with/without the morphological analyzer reinforces its role. Even with a better model, morphological analyzers still provide additional consistency between the different features. We will provide additional details at the error analysis section, but further research is required to enhance the consistency of the generated features, without needing a morphological analyzer.

\paragraph{\textbf{\textsc{BlindTest}} Results}

The results for the \textsc{BlindTest} dataset were consistent with the \textsc{DevTest}. The accuracy for \textsc{Egy} using the strongest baseline is  78.1, based on the multitask learning architecture for the tags. The accuracy of the best system, using the joint modeling architecture along with the morphological analyzer, is 80.3. We also observed the same behavior for MSA, with somewhat similar values to  \textsc{DevTest}.

\subsection{Error Analysis}

\paragraph{The Role of Morphological Analyzers} The goal is to assess the role of morphological analyzers in the consistency (following the consistency definition mentioned earlier) of the predicted features. We took a sample of 1000 words from the MSA \textsc{DevTest}, and ran it through the joint model that does not use a morphological analyzer, and checked the errors in the predictions. There were 110 errors (11\% of the sample), for an accuracy of 89\%, which is close to the reported accuracy over the entire dataset. About 62\% of the errors had consistent feature predictions, but the predicted analysis did not match the gold. And around 13\% of the errors are due to gold errors.

Around 25\% of the errors (2.8\% of the sample) had inconsistent predictions. This roughly matches the accuracy gap between the joint model with and without the morphological analyzer, which is also around 2\%. This indicates that the accuracy boost that the morphological analyzer provides is to a large extent due to the consistency it conveys.  We also observed that 37\% of the inconsistent predictions (1\% of the sample) had a correct lemma, but the lemma was inconsistent with the rest of the features. The remaining 63\% (1.7\% of the sample), had an invalid lemma.

\paragraph{Joint Modeling vs Separate Modeling} We also investigated the distribution of errors over the different features for the joint model against the baseline of separate feature models, both using the morphological analyzer. We annotated the errors in a 1000-word sample from \textsc{DevTest}, for both MSA and \textsc{Egy}, with the main erroneous feature. For example, if the predicted analysis is a verb inflection of a gold noun, the main erroneous feature would be the POS tag, even if other features ended up being wrong as a result. For MSA, the error distribution for the baseline is:  case 27\%, diacritization 22\%, POS 18\%, lemmatization 13\%, gold errors 11\%, and smaller percentages for state, voice, person, and enclitics.
Whereas the distribution for the joint model is: case 26\%, POS 21\%, lemmatization 18\%, gold errors 14\%, diacritization 13\%, and small percentages for state, voice, and person. In both models, case  dominates the error distribution, since identifying the case ending in MSA is particularly challenging. The main difference between the models in terms of error distribution is the diacritization, where we observe a significant boost when we use the joint model. The apparent increase in the error percentages of the other error types at the joint model is due to the drop in the overall errors count, while many of these error types seem to have a lower drop rate.

For \textsc{Egy}, a notable error pattern is when the prediction matches the MSA-equivalent analysis of the dialectal word, like having an MSA-like diacritization, or having a case ending (DA, like \textsc{Egy}, does not have case ending). This happens due to code-switching with MSA in the dialectal content in general, which is also reflected at the analyzer. This error type is not an error per se, but we do include it in the analysis. The error distribution for the separate features baseline is: gold errors 23\%, MSA equivalents 21\%, POS 17\%, lemmatization 14\%, diacritization 12\%, and smaller percentages for several other error types. Whereas the distribution for the joint model is: gold errors 27\%, MSA equivalents 21\%, lemmatization 18\%, POS 14\%, diacritization 7\%, and smaller frequencies for the other errors. The amount of gold errors is significant, but it is consistent with other contributions that use the same dataset \cite{Zalmout_2018_NAACL}. 

Similar to MSA, the increase in the error percentages of the other error types at the joint model is due to the drop in the overall errors count, while the other error types seem to have a lower drop rate, especially for the MSA equivalents and gold errors, which are inherent in the dataset, and will not be affected by better modeling.

\section{Conclusions and Future Work}
We presented a joint modeling approach for the lexicalized and non-lexicalized features in morphologically rich and Semitic languages. Our model achieves a significant improvement over several baselines for Arabic, and matches the baseline for MSA without having to use an expensive morphological analyzer. The results highlight the benefits of joint modeling, where diacritization seems to have benefitted the most. 
We observed, however, that further research is needed to enhance the overall consistency of the predicted features, without having to rely on external morphological analyzers.

 \paragraph{Acknowledgment}
The first author was supported by the New York University Abu Dhabi Global PhD Student Fellowship program. The support and resources from the High Performance Computing Center at New York University Abu Dhabi are also gratefully acknowledged.

\bibliographystyle{acl_natbib}

\end{document}